\documentclass[sigconf,nonacm]{acmart}

\renewcommand\footnotetextcopyrightpermission[1]{}
\usepackage{dirtree}
\usepackage{float}
\usepackage{algorithm}
\usepackage{algpseudocode}

\begin{document}

\title{Who Belongs in the Eval Set? \\ A Capability-Taxonomy-Driven Pipeline for Curating Regression Eval Sets in Agent-Extensibility Platforms}

\author{Tezan Sahu}
\affiliation{%
  \institution{Microsoft}
  \city{Hyderabad}
  \country{India}
}
\email{tezansahu@microsoft.com}

\author{Aritra Das}
\authornote{Work performed during an internship at Microsoft.}
\affiliation{%
  \institution{Indian Institute of Technology Roorkee}
  \city{Roorkee}
  \country{India}
}
\email{aritra_d@cs.iitr.ac.in}

\author{Pankaj Mittal}
\affiliation{%
  \institution{Microsoft}
  \city{Hyderabad}
  \country{India}
}
\email{pankajmi@microsoft.com}

\author{Sudipta Das}
\affiliation{%
  \institution{Microsoft}
  \city{Hyderabad}
  \country{India}
}
\email{sudiptad@microsoft.com}

\begin{abstract}
Platform teams hosting agent-extensibility surfaces face a regression-economics paradox: every onboarding customer ships an evaluation set tuned to their domain, but the platform's regression set must live under a hard query-count ceiling structurally bounded by release cadence. \textbf{To the best of our knowledge, no published industrial pipeline addresses this platform-side curation problem} --- existing evaluation frameworks~\cite{helm,agentbench,hal,langsmith,braintrust,foundryevals} are customer-side, and benchmark-compression work~\cite{essencebench,metabench,tinybench} treats benchmarks as fixed pools rather than streams of incoming sets. We describe a capability-taxonomy-driven curation pipeline being applied to declarative agents with custom actions in Microsoft 365 Copilot~\cite{copilotextend}. It takes an agent specification and a customer's eval set as input, projects each query into a platform-owned \emph{capability taxonomy} ($\sim$30 capabilities), and outputs per-query decisions (admit, drop, swap, or human review) under the philosophy that a healthy regression set is the \emph{minimal collection of queries that captures the maximal spread of capability signatures} --- distinct combinations of capabilities a query exercises together. Three components instantiate this: a \emph{classifier} producing per-(query, capability) verdicts via a hybrid of deterministic specification-based extraction and large-language-model (LLM) semantic inference~\cite{wang2023selfcons}; an \emph{Invocation Quality (IQ) rater} that scores how thoroughly a query exercises each capability, so a new query sharing a signature with an existing entry can still be recognized as a better test and displace it; and a \emph{consolidator} that compares incoming queries against the regression set on coverage \emph{and} quality through a rule-based decision cascade, backed by a conservative curator that only suggests evictions. The mechanism is taxonomy-agnostic and applies to any regression eval-set curation problem with a typed capability taxonomy --- including ones whose taxonomies evolve in response to the very evidence the pipeline surfaces.
\end{abstract}

\maketitle

\begin{figure*}[t]
\centering
\includegraphics[width=0.95\textwidth]{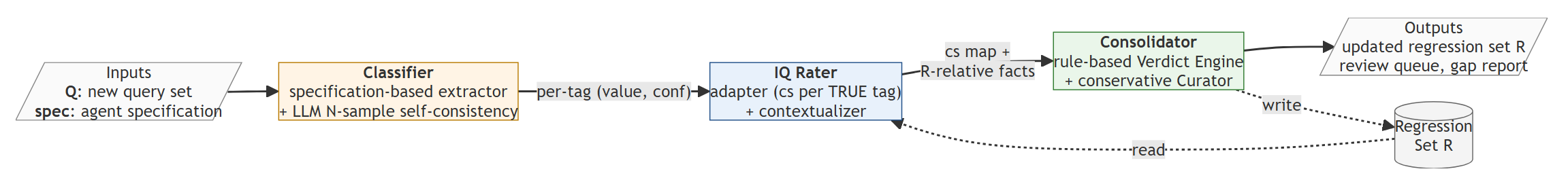}
\caption{Pipeline overview. The classifier emits per-tag verdicts and confidences for each (query, capability) pair. The Invocation Quality (IQ) rater turns these into a uniform per-capability score and derives $R$-relative facts (champions, new coverage, wins, similarity, novelty). The consolidator's pure-rule Verdict Engine emits one decision per query; the Curator is the only writer to the regression set $R$.}
\label{fig:pipeline}
\end{figure*}

\section{Introduction}

Agent-extensibility platforms expose a menu of composable capabilities --- authentication modes, tool-invocation protocols (REST/API-based plugins, Model Context Protocol~\cite{mcp}, MCP Apps~\cite{mcpapps}), orchestration patterns (single-turn, multi-turn, parallel calls, branching), and response-format primitives --- that downstream third parties compose into agents. Every onboarding customer brings an evaluation set tuned to their domain.

\textbf{Why customer eval sets, not synthetic ones?} A reasonable alternative is for the platform to synthesize its own eval set by enumerating capability combinations. Two reasons rule this out. \emph{Combinatorial blowup}: with $\sim$30 typed capabilities the joint configuration space is intractable, and uniform sampling produces configurations no real customer would ever build. \emph{LLM non-determinism over realistic phrasings}: per-capability behavior is verified by unit and functional tests we run separately, but end-to-end agent behavior --- multiple capabilities firing through an LLM under a real user's wording --- is governed by how the LLM responds to actual customer phrasings, a distribution synthetic queries cannot reproduce. Customer eval sets carry both the combinations real agents ship and the phrasings the LLM has to handle. Two structural pressures then bound this setting:

\textbf{Hard eval-set ceiling.} The regression set is judged by a large language model (LLM) on every candidate release build and human-triaged on every regression failure, gating flight reviews, A/B tests, and release certification. Every admission competes for a finite slot bounded by release cadence.

\textbf{Tenant-and-mock-setup labor.} Regression evals must \emph{deterministically} exercise the platform's code paths, including those touching third-party services --- hitting real 3P endpoints is unviable, since they are flaky, rate-limited, costly, and behaviorally non-stationary. Each admitted query therefore carries a provisioned tenant, synthetic connectors, and \emph{mock data fixtures} engineered to deterministically reproduce the expected 3P responses. This per-query reviewer labor often dominates the release's token-and-compute cost.

The naive ``admit all of your queries'' inflates this engineering tail; the equally naive ``manual review committee'' freezes intuition and degrades with drift. Two questions must be answered per incoming query: \textbf{(Q1)} which capabilities does it exercise? \textbf{(Q2)} given the live regression set, does it deserve a slot?

Existing literature offers neither answer in this form. Customer-side LLM and agent evaluation frameworks --- HELM~\cite{helm}, AgentBench~\cite{agentbench}, HAL~\cite{hal}, together with observability platforms like LangSmith~\cite{langsmith}, Braintrust~\cite{braintrust}, and Azure AI Foundry Evaluation~\cite{foundryevals} --- certify a \emph{single} agent owner against their own success criteria; the notion of a platform-owned regression set curated across many customers under one fixed budget is simply not their setting. Benchmark-compression methods~\cite{essencebench,metabench,tinybench} achieve up to $200\times$ pool reductions while preserving ranking, but operate on closed, fixed pools rather than an open stream feeding a long-lived, edit-in-place set. Test-suite minimization in software engineering~\cite{chvatal1979,smithtsm,ltm} supplies a set-cover framing on the coverage axis (NP-complete; greedy yields an $H_n$-approximation), but classical formulations carry no per-capability \emph{quality} dimension --- the very dimension a same-signature replacement decision needs. \textbf{To the best of our knowledge, no published industrial pipeline addresses this platform-side eval-set curation problem.} What we describe is therefore not a one-shot compression of a fixed pool but the curation policy for a \emph{living regression set} that ingests new customer eval sets continuously.

We close that gap. The rest of the paper develops a three-component pipeline --- a hybrid classifier, an Invocation Quality (IQ) rater, and a rule-based consolidator with a conservative curator --- that jointly answers Q1 and Q2 over a platform-owned capability taxonomy, and we are applying it to declarative agents with custom actions in Microsoft 365 Copilot~\cite{copilotextend}.

\section{Methodology}

The three components are chained in sequence (Fig.~\ref{fig:pipeline}) over a platform-owned capability taxonomy and a live regression set $R$. Each component below first defines its outputs and the reason it exists; the design invariants underpinning the whole are summarized in \S\ref{sec:invariants}.

\subsection{Capability Taxonomy and Signatures}\label{sec:taxonomy}

Let $\mathcal{C} = \{c_1, \ldots, c_m\}$ be the platform's capability taxonomy --- a typed, hierarchical classification (the deployed instance has $m \approx 30$ leaves). Mutually-exclusive groups carry \texttt{exactly-one} constraints. A slice (compact form, sibling lists wrap by line):

{\footnotesize
\dirtree{%
.1 Capability Taxonomy.
.2 Request.
.3 Auth \{No Auth, API Key, OAuth, SSO\}.
.3 Attachments \{Image, File, Audio\}.
.2 Execution.
.3 Orchestration \{single/multi-turn, parallel, branching\}.
.3 Tool Output Size \{Small, Normal, Large\}.
.2 Response.
.3 Format \{Plain, Structured, URLs, Tables\}.
}
}

For a query $q$, the classifier emits a value $v_{q,c} \in \{\mathrm{TRUE}, \mathrm{FALSE}\}$ for every $c \in \mathcal{C}$. We call $\mathrm{sig}(q) = \{c : v_{q,c} = \mathrm{TRUE}\}$ the \emph{capability signature} of $q$. The regression set $R$ is the curated portfolio of queries; the pipeline's job is to keep $R$ small while $\bigcup_{q \in R} \mathrm{sig}(q)$ stays wide.

The taxonomy is itself versioned. Platforms add capabilities as features ship, and reviewer feedback may also \emph{retro-add} dimensions: when many incoming queries land outside $\mathcal{C}$ (caught as low-confidence, off-distribution clusters at projection time), the pipeline surfaces these as taxonomy-extension candidates --- either truly emergent platform usage, capabilities the original taxonomy missed, or new combinations worth naming. Reviewer-approved extensions bump the taxonomy version and re-classify affected queries.

\subsection{Classifier}\label{sec:classifier}

For each $(q, c)$ the classifier emits a verdict $v_{q,c}$ and a confidence $\kappa_{q,c} \in [0,1]$ via two non-overlapping paths.

\textbf{Specification-based extractor.} For capabilities the agent specification declares (authentication mode, tool inventory, declared platform capabilities), the truth is a lookup over the spec. The extractor returns the declared value with $\kappa_{q,c} = 1$.

\textbf{LLM with $N$-sample self-consistency~\cite{wang2023selfcons}.} For capabilities that surface only in query intent (response-format expectations, branching logic, tonality), the classifier draws $N$ independent LLM samples, each returning a boolean vote $\tilde v_i$ and an exercise-probability $P_i$. Then:
\[
v_{q,c} = \mathrm{majority}(\tilde v_1, \ldots, \tilde v_N), \quad \kappa_{q,c} = \tfrac{1}{N}\textstyle\sum_{i=1}^{N} P_i.
\]
The two paths are kept non-overlapping --- the specification-based path is authoritative where the spec speaks; inference never overrides ground truth. Mutex arbitration over \texttt{exactly-one} groups runs after the paths resolve.

\textbf{Evaluation.} We evaluate the LLM path (the specification-based path is a deterministic lookup) by \emph{per-capability correlation between predicted verdicts and human-labeled ground truth} rather than raw accuracy. The choice is forced by class imbalance: each query exercises only a handful of the $\sim$30 capabilities, so an always-FALSE baseline would score $>$90\% accuracy by construction --- an uninformative number. Per-capability correlation against labels captures the agreement structure correctly under this imbalance and avoids being inflated by the dominant negative class.

\subsection{Invocation Quality (IQ) Rater}\label{sec:rater}

The rater answers what binary coverage cannot: when a new query shares $\mathrm{sig}(q)$ with an existing $r \in R$, is $q$ a \emph{better} test of those capabilities than $r$? Overlap is common; a shallow tester should yield to a deeper one. The rater splits into two halves --- an \emph{intrinsic} adapter (a fact about $q$ alone) and a \emph{relational} contextualizer (a fact about $q$ versus the live $R$). The split is load-bearing: only the intrinsic half can be persisted; the relational half must be recomputed every cycle, because $R$ itself changes on every admission and eviction.

\textbf{Adapter (intrinsic, stateless).} For every $c$ with $v_{q,c} = \mathrm{TRUE}$, the adapter emits
\[
\mathrm{cs}(q, c) := \kappa_{q,c} \in [0, 1],
\]
the answer to ``how thoroughly does $q$ exercise $c$?''. $\mathrm{cs}$ is a stored fact about $q$ alone --- a property of the query, not of the regression set.

\textbf{Contextualizer (relational, live).} A pure intrinsic score is not enough to decide what to do with $q$: the consolidator's rules need to know whether $q$ \emph{fills a coverage gap}, \emph{decisively beats an incumbent}, or is just a \emph{near-duplicate} of something we already keep. None of those questions can be answered from $\mathrm{cs}(q, \cdot)$ in isolation --- every one of them refers to what is currently in $R$. The contextualizer is the bridge: it materializes, for $q$ against the live $R$, six relational facts that encode exactly these questions. Two calibration knobs enter here: $\tau$ is the \emph{competence bar} (the minimum $\mathrm{cs}$ for a capability to count as competently covered), and $\delta$ is the \emph{winning margin} (by how much an incoming score must beat the incumbent champion to count as decisive). Both live in configuration (\S\ref{sec:invariants}, I4). Because $R$ changes every cycle, the facts below are derived fresh and never persisted (\S\ref{sec:invariants}, I2):
\[
\begin{aligned}
&\mathrm{champion}(c) = {\arg\max}_{r \in R}\, \mathrm{cs}(r, c), &&\text{\footnotesize the incumbent best test of $c$;} \\
&\mathrm{champ\_cs}(c) = {\max}_{r \in R}\, \mathrm{cs}(r, c), &&\text{\footnotesize current bar to beat on $c$;} \\
&\mathrm{caps}_\tau(q) = \{c : \mathrm{cs}(q, c) \geq \tau\}, &&\text{\footnotesize $q$'s \emph{competent} coverage} \\[-2pt]
& &&\text{\footnotesize (effective, not nominal);} \\
&\mathrm{newCaps}(q) = \mathrm{caps}_\tau(q) \setminus {\bigcup}_{r \in R} \mathrm{caps}_\tau(r), &&\text{\footnotesize ``$q$ fills a gap''} \\[-2pt]
& &&\text{\footnotesize (no one in $R$ covers these well);} \\
&\mathrm{wins}(q) = \{c : \mathrm{cs}(q, c) > \mathrm{champ\_cs}(c) + \delta\}, &&\text{\footnotesize ``$q$ beats the champion''} \\[-2pt]
& &&\text{\footnotesize (by margin $\delta$);} \\
&\mathrm{maxSim}(q) = {\max}_{r \in R}\, \mathrm{sim}(q, r), &&\text{\footnotesize clone-detection signal;} \\
&\mathrm{novel}(q) = \mathbb{1}\!\left[\mathrm{sig}(q) \notin \{\mathrm{sig}(r) : r \in R\}\right]. &&\text{\footnotesize unseen capability combination.}
\end{aligned}
\]
These six facts are the only inputs the Verdict Engine consumes about $R$; the engine itself touches no $\mathrm{cs}$ values directly.

\textbf{Evaluation.} Absolute quality scores for individual queries are not available in the abstract; what \emph{is} available, by design, is reviewer action on the verdicts the pipeline emits. We therefore evaluate the rater at the verdict level: every admission, replacement, and eviction-candidate surfaced by the consolidator passes through a reviewer in the normal release flow, and their accept/reject call is logged. Verdict-level agreement between the pipeline and the reviewer is the quality metric. This reuses the review signal already produced for production purposes, so quality measurement imposes no additional dedicated annotation cycles.

\subsection{Consolidator}\label{sec:consolidator}

Given $\mathrm{cs}(q, \cdot)$ and the $R$-relative facts, a pure rule-based Verdict Engine emits one of $\{\mathrm{INCLUDE}, \mathrm{EXCLUDE}, \mathrm{REPLACE}, \mathrm{KEEP\text{-}BOTH}\}$, each tagged \emph{direct} or \emph{review}. Four further calibration knobs enter at this layer: $\sigma$ is the \emph{clone-similarity cutoff} (text similarity above which an incoming $q$ is treated as a near-duplicate of some $r \in R$); $N$ is the \emph{redundancy floor} (minimum number of competent coverers $R$ must retain per capability before any one of them can be evicted); $k$ is the \emph{concentration cap} (limit on how many capabilities a single $r$ may be the \emph{sole} competent coverer of); $\mathrm{cs}_{\min}$ is the \emph{honesty floor} (minimum top $\mathrm{cs}$ across $q$'s TRUE tags below which a direct-admission verdict is downgraded to review). With these in hand, the cascade fires five gates in fixed order --- (1)~clone gate $\mathrm{maxSim}(q) \ge \sigma$; (2)~merit ($\mathrm{newCaps} \cup \mathrm{wins} \cup \{c: \mathrm{novel}(q)\} \neq \emptyset$); (3)~new competent coverage ($\mathrm{newCaps} \neq \emptyset$); (4)~replacement eligibility (subset-domination, $N$, $k$); and (5)~honesty downgrade (admitting on direct route with $\max_c \mathrm{cs}(q, c) < \mathrm{cs}_{\min}$ forces \emph{review}). Each gate contributes at most one reason code; the full pseudocode is in Appendix~\ref{app:rules}.

\textbf{Curator.} The only writer to $R$. It stages admissions per query, prunes per cycle, and \emph{never auto-deletes}: redundant un-pinned coverers are flagged as suggest-only eviction items that a human confirms. The asymmetry between the pure Verdict Engine and the write-capable Curator is the core safety property (\S\ref{sec:invariants}, I3).

\subsection{Design Invariants}\label{sec:invariants}

The pipeline rests on five invariants. With the components in hand, each becomes a constraint on the math we have just defined.

\textbf{I1: Atomicity.} A query is the smallest curatable unit. If it is the best test of \emph{any one} capability (its $\mathrm{cs}$ is the champion on that capability and clears $\tau$), it is \emph{pinned} and cannot be evicted; we tolerate extra redundancy elsewhere instead.

\textbf{I2: Stored facts, derived roles.} $\mathrm{cs}(q, c)$ is a stored fact about $q$ alone. The $R$-relative roles --- \emph{champion}, \emph{redundant}, \emph{pinned}, \emph{covered} --- are re-derived live whenever $R$ changes. The intrinsic-$\mathrm{cs}$ / $R$-relative-role line \emph{is} the rater/consolidator boundary.

\textbf{I3: Admission cheap, eviction slow.} A REPLACE verdict never deletes; it stages a candidate. The Curator's prune turns a beaten resident into a \emph{suggested} eviction item that a human confirms. Eviction removes evidence the platform was relying on, and must remain auditable.

\textbf{I4: No hard thresholds in code.} All six knobs introduced above ($\tau$, $\delta$, $\sigma$, $N$, $k$, $\mathrm{cs}_{\min}$) live in configuration, never in code. Their current calibration and the open problem of moving from uniform defaults to per-capability values are discussed in \S\ref{sec:discussion}.

\textbf{I5: Honesty as route-downgrade, not admission.} A low-confidence tag cannot \emph{admit} a query; if rules yield an admitting verdict whose value rests on a low-confidence tag, the route is forced to review with reason \texttt{LOW\_CONFIDENCE}. An EXCLUDE is never converted into an admission.

\section{Discussion}\label{sec:discussion}

We are applying this pipeline to the curation of regression eval sets for declarative agents with custom actions in Microsoft 365 Copilot. Two threads stand out.

\textbf{Generalizability, with an evolving taxonomy.} The mechanism is \emph{taxonomy-agnostic}: the classifier, IQ rater, and consolidator make no assumption about which capabilities the taxonomy contains, only that it is typed and supports \texttt{exactly-one} groups. The same pipeline applies to any regression-eval-set curation problem where (a) a platform team owns a capability taxonomy, (b) downstream contributors ship eval sets, and (c) the curated set is \emph{living} and bounded by a fixed budget. Crucially, the taxonomy itself is not fixed: platforms add capabilities over time, and the pipeline closes a loop in which queries that project to low-confidence, off-distribution regions cluster into taxonomy-extension candidates that reviewers can promote into the next taxonomy version (whether they correspond to emergent platform behavior, missed combinations, or capabilities the original taxonomy did not anticipate). Web service APIs, data pipelines, and other multi-tenant platforms with evolving capability menus are natural next targets.

\textbf{Per-capability calibration is the central tuning problem.} The six knobs $\tau$, $\delta$, $\sigma$, $N$, $k$, $\mathrm{cs}_{\min}$ are absolute thresholds, not batch-relative percentiles, and a value sensible for one capability is not automatically sensible for another --- the bar for ``competent coverage'' of authentication is structurally different from the bar for response-format handling. The principled target is therefore a per-capability schedule of these knobs, learned from validated reviewer overrides. Our current workaround is a uniform, conservative default set across all capabilities, derived from a small pilot of reviewer-annotated runs; reviewer overrides on borderline verdicts are logged but do not yet feed an automated calibration loop. Identifying which override signal counts as ``validated'' --- separating principled adjustments from reviewer disagreement noise --- is the next architectural decision.

\textbf{Pipeline accessibility via an MCP server.} We expose the pipeline behind a Model Context Protocol server, with tools mapped to each pipeline step (classify, rate, consolidate) and to common step-ranges (classify-to-verdict; $\mathrm{cs}$-to-verdict, for re-running the consolidator over an updated $R$). This lets any MCP-aware coding agent or chat assistant invoke the pipeline programmatically: reviewers triage incoming eval sets from their existing chat surface, and partner platforms can embed the pipeline into their own onboarding flows without re-implementing it. Combined with the taxonomy-agnostic design above, this is what makes the system reusable rather than Microsoft-365-specific.

\section{Conclusion \& Future Work}

We presented a capability-taxonomy-driven pipeline for the platform-side curation of regression eval sets, an industrial problem we believe is unaddressed in the published literature. Three components --- a hybrid classifier, an intrinsic-versus-relative IQ rater, and a rule-based consolidator with a conservative curator --- jointly maintain a small regression set with maximal capability-signature coverage on top of a versioned, evolving taxonomy. Future work spans per-capability threshold calibration, recency-decayed champion selection so stale-but-strong incumbents yield to fresher tests, cost-weighted set-cover when capabilities differ materially in setup labor, and an end-to-end study of how the pipeline transfers to non-agent platforms.

\bibliographystyle{ACM-Reference-Format}

\appendix
\section*{Appendix}
\renewcommand{\thesection}{\Alph{section}}
\setcounter{section}{0}

\section{Worked Example}\label{app:example}

A customer onboards a project-tracker agent (OAuth-authenticated, four read-type tools, one write-type tool) and ships a 20-query eval set with their onboarding bundle. At the start of this cycle, the regression set $R$ already holds $\sim$120 queries accumulated across prior onboardings of similar agents: several entries exercise OAuth-gated read flows, but no incumbent strongly tests structured-output formatting (the best Structured Output coverer in $R$ currently sits at $\mathrm{cs}=0.30$, below the competence bar $\tau=0.50$). We walk one of the 20 incoming queries through the pipeline:
\begin{quote}\itshape ``Pull recent issues from the tracker filtered by my team's tag and summarise them as a bulleted list.''\end{quote}
Tables~\ref{tab:example-classifier} and~\ref{tab:example-iq} give the classifier output and the resulting $R$-relative facts for this query.

\begin{table}[H]
\footnotesize
\centering
\setlength{\abovecaptionskip}{1pt}
\setlength{\belowcaptionskip}{1pt}
\caption{Classifier output (TRUE tags only).}
\label{tab:example-classifier}
\begin{tabular}{@{}lll@{}}
\toprule
\textbf{Capability} & \textbf{Path} & \textbf{$\kappa$} \\
\midrule
OAuth & spec-extractor & 1.00 \\
Has Read Tools & spec-extractor & 1.00 \\
Single-turn & LLM ($N{=}5$) & 0.72 \\
Structured Output & LLM ($N{=}5$) & 0.45 \\
\bottomrule
\end{tabular}
\end{table}

\vspace{-8pt}

\begin{table}[H]
\footnotesize
\centering
\setlength{\abovecaptionskip}{1pt}
\setlength{\belowcaptionskip}{1pt}
\caption{$R$-relative facts for $q$ ($\tau=0.50$, $\sigma=0.80$, $\delta=0.05$). The Structured Output champion in $R$ has $\mathrm{champ\_cs}=0.30 < \tau$ --- below the competence bar.}
\label{tab:example-iq}
\begin{tabular}{@{}p{2.6cm}l@{}}
\toprule
\textbf{Fact} & \textbf{Value} \\
\midrule
$\mathrm{caps}_\tau(q)$ & \{OAuth, Has Read Tools, Single-turn\} \\
$\mathrm{newCaps}(q)$ & \{Structured Output\} \\
$\mathrm{wins}(q)$ & $\emptyset$ \\
$\mathrm{maxSim}(q)$ & 0.42 \\
$\mathrm{novel}(q)$ & 0 \\
\bottomrule
\end{tabular}
\end{table}

\vspace{-4pt}

\noindent\textbf{Verdict.} Clone gate passes ($0.42 < 0.80$), merit holds ($\mathrm{newCaps}$ non-empty), and new competent coverage is non-empty, so the engine emits \textbf{INCLUDE-direct} with reason \texttt{NEW\_COVERAGE}; the honesty check ($\max_c \mathrm{cs}(q, c) = 1.0 \ge \mathrm{cs}_{\min}$) does not downgrade. \textbf{Curator action.} Stages $q$ into $R$, and on the next per-cycle prune the previous shallow Structured Output coverer ($\mathrm{cs}=0.30$) is dethroned and surfaced as a suggest-only eviction item that a human confirms before deletion.

\section{Verdict Engine: Formal Rules}\label{app:rules}

\begin{algorithm}[H]
\footnotesize
\caption{Verdict Engine}
\begin{algorithmic}[1]
\Function{Decide}{$q$, $\mathrm{RFacts}$, $\mathrm{cfg}$}
  \If{$\mathrm{maxSim}(q) \ge \sigma$} \Return $(\textsc{Exclude}, \mathrm{direct}, \textsc{Clone})$
  \EndIf
  \State $\mathrm{merit} \gets (\mathrm{newCaps} \neq \emptyset) \lor (\mathrm{wins} \neq \emptyset) \lor \mathrm{novel}(q)$
  \If{$\neg\, \mathrm{merit}$} \Return $(\textsc{Exclude}, \mathrm{direct}, \textsc{Redundant})$
  \EndIf
  \If{$\mathrm{newCaps} \neq \emptyset$}
    \State $\mathit{verdict} \gets (\textsc{Include}, \mathrm{direct}, \textsc{New\_Coverage})$
  \ElsIf{$\mathrm{wins} \neq \emptyset$}
    \State $E \gets \{\mathrm{champion}(c) : c \in \mathrm{wins}\}$ \Comment{evict candidates}
    \If{$\exists r \in E : \mathrm{caps}(r) \not\subseteq \mathrm{caps}(q)$}
      \State \Return $(\textsc{Keep-Both}, \mathrm{direct}, \textsc{Split\_or\_Floor})$
    \ElsIf{$\exists c : \#\mathrm{coverers}_{R'}(c) < N$ on $R' = (R \setminus E) \cup \{q\}$}
      \State \Return $(\textsc{Keep-Both}, \mathrm{direct}, \textsc{Split\_or\_Floor})$
    \ElsIf{$\mathrm{sole}(q) > k$}
      \State \Return $(\textsc{Replace}, \mathrm{review}, \textsc{Concentration})$
    \Else
      \State $\mathit{verdict} \gets (\textsc{Replace}, \mathrm{direct}, \textsc{Dominates})$
    \EndIf
  \Else
    \State $\mathit{verdict} \gets (\textsc{Include}, \mathrm{review}, \textsc{Novel\_Combo})$
  \EndIf
  \If{admitting($\mathit{verdict}$) $\land$ route $=$ direct $\land \max_c \mathrm{cs}(q, c) < \mathrm{cs}_{\min}$}
    \State $\mathit{verdict}$.route $\gets$ review; append \textsc{Low\_Confidence}
  \EndIf
  \State \Return $\mathit{verdict}$
\EndFunction
\end{algorithmic}
\end{algorithm}

\end{document}